 \documentclass[final,5p,times,twocolumn]{elsarticle}

\journal{Pattern Recognition Letters}

\usepackage{graphicx}
\usepackage{epsfig,times}
\usepackage{array}
\usepackage{multirow}
\usepackage{subfigure}
\usepackage{xcolor}
\usepackage{enumerate}
\usepackage{hyperref}
\usepackage{algorithm,algorithmic}
\usepackage{amsmath,amssymb,amsfonts}
\usepackage{bm}
\graphicspath{{image/}} 

\newenvironment{proof}[1][Proof]{\begin{trivlist}
\item[\hskip \labelsep {\bfseries #1}]}{\end{trivlist}}

\renewcommand{\d}{\mathbf{d}}
\newcommand{\x}{\mathbf{x}}
\newcommand{\balpha}{{\bm{\alpha}}}
\newcommand{\bmu}{{\bm{\mu}}}
\renewcommand{\r}{\mathbf{r}}
\newcommand{\y}{\mathbf{y}}
\newcommand{\e}{\mathbf{e}}
\newcommand{\z}{\mathbf{z}}

\newcommand{\by}{\mathbf{Y}}
\newcommand{\bd}{\mathbf{D}}

\def\etal{\emph{et al.}}
\def\ie{\emph{i.e.}}
\def\eg{\emph{e.g.}}
\begin{document}

\begin{frontmatter}


\title{Sparse Coding and Counting for Robust Visual Tracking}

\author[2]{Risheng Liu\corref{cor1}}
\author[1]{Jing Wang}
\cortext[cor1]{Corresponding author:
  Tel.: +86-411-84708300 87571640;
  fax: +86-411-84671713;}
\ead{rsliu@dlut.edu.cn}
\author[1]{Yiyang Wang}
\author[1]{Zhixun Su}
\author[1]{Yu Cai}
\address[1]{School of Mathematic Sciences, Dalian University of Technology, China}
\address[2]{School of Software Technology, Dalian University of Technology, China}
\address[3]{Key Laboratory for Ubiquitous Network and Service Software of Liaoning Province, China}


\begin{abstract}
In this paper, we propose a novel sparse coding and counting method under Bayesian framwork for visual tracking.
In contrast to existing methods, the proposed method employs the combination of $L_0$ and $L_1$ norm
to regularize the linear coefficients of incrementally updated linear basis.
%JWL{we are not using $L_0$ to represent the appearance, but to regularize the linear coefficients.}
%to represent the object and select the most informative candidate for better prediction.
%
The sparsity constraint enables the tracker to effectively handle difficult challenges, such as occlusion or image corruption.
%Moreover, some challenging issues, such as occlusion, corruption, varying illumination,
%are also addressed seamlessly through the sparsity constraint.
%
To achieve realtime processing, we propose
%We also aim at developing
a fast and efficient numerical algorithm for solving the proposed model.
%
%Although the problem involving $L_0$ norm is an NP-hard problem,
%we develop a fast and efficient algorithm based on the accelerated
%proximal gradient (APG) approach with guaranteed convergency.
Although it is an NP-hard problem, the proposed accelerated proximal gradient (APG) approach is guaranteed to converge to a solution quickly.
Besides, we provide a closed solution of combining $L_0$ and $L_1$ regularized representation to obtain better sparsity.
Experimental results on challenging video sequences
demonstrate that the proposed method achieves state-of-the-art results
both in accuracy and speed.

\end{abstract}


\end{frontmatter}

% \linenumbers


\section{Introduction}
\label{sect:intro}  % \label{} allows reference to this section
Visual tracking plays an important role in computer vision and has many applications such as video surveillance, robotics, motion analysis and human computer interaction.
Even though various algorithms have come out, it is still a challenge problem due to complex object motion, heavy occlusion, illumination change and background clutter.
%It is a challenging work due to the difficulties to account for appearance variation of a target object caused by intrinsic pose variation, shape deformation and extrinsic illumination change, camera motion, occlusion.
%Despite great progresses in the past decades, due to numerous challenges in real world, many challenging problems still remain when designing a practical visual tracking system.
%%
%For example, pose variation, shape deformation, varying illumination,
%camera motion, and occlusions may increase the difficulty for
%visual tracking algorithms.
%
%In recent years, there have been extensive literatures on object tracking.
%

Visual tracking algorithms can be roughly categorized into two major categories:
discriminative methods and generative methods.
Discriminative methods (\eg,~\cite{co/training/iccv/LiuCL09,MIL/cvpr/BabenkoYB09,Struck/iccv/HareST11}) view object tracking as a binary classification problem in which the goal is to separate the target object from the background.
Generative methods (\eg,~\cite{Appearance/Models/pami/JepsonFE03,IVT/ijcv/RossLLY08,L1PCA/Liu14,LRF/Zhang14,LSP/Liu14})
employ a generative appearance model to represent the target's appearance.

%We focus on developing a robust generative tracking method, which can successfully model the target observations and deal with difficult challenges using novel sparse regularized representation.
We focus on the generative one and will briefly review the relevant work below.
Recently, sparse representation has been successfully applied to visual tracking (\eg,~\cite{l1/tracker/mei,Liu/sparse/represnetation/tracking/application,MTT/ijcv/ZhangGLA13,conf/icmcs/Jinwei}).
The trackers based on sparse representation are under the assumption that the appearance of a tracked object can be sparsely represented by a over-complete dictionary which can be dynamically
updated to maintain holistic appearance information. Traditionally, the over-complete dictionary is a series of redundant object templates, however, a set of basis vectors from target subspace as dictionary is also used because an orthogonal dictionary performs as efficient
as the redundant one. In visual tracking, we will call the $L_1$ regularized object representation "sparse coding" (\eg,~\cite{l1/tracker/mei}), and the $L_0$ regularized object representation "sparse counting" (\eg,~\cite{panl0}).
\cite{l1/tracker/mei} has been shown to be robust against partial occlusions,
which improves the tracking performance. However, because of using redundant dictionary, heavy computational overhead in $L_1$ minimization hampers the tracking speed.
%sparse coding based trackers perform computationally
%expensive $L_1$ minimization at each frame.
Very recent efforts have been made to improve this method in
terms of both speed and accuracy by using accelerated proximal gradient (APG) algorithm~\cite{APGL1/bao}
or modeling the similarity between different candidates~\cite{MTT/ijcv/ZhangGLA13}.
Different from~\cite{l1/tracker/mei}, IVT~\cite{IVT/ijcv/RossLLY08} incrementally learns a low-dimensional PCA subspace representation, which adapts online to the appearance changes of the target.
%the works in~\cite{IVT/ijcv/RossLLY08,Lu/tip13/Wang13,panl0} point out that the aforementioned
%method do not exploit rich and redundant image properties which can be captured
%compactly with subspace representations.
%
To get rid of image noise, Lu \etal~\cite{Lu/tip13/Wang13} introduce $L_1$ noise regularization into the PCA reconstruction, which is able to handle partial occlusion and other challenging factors.
Pan \etal~\cite{panl0} employs $L_0$ norm to regularize the linear coefficients of incrementally updated linear basis (sparse counting) to remove the redundant features of the basis vectors.
However, sparse counting will cause unstable solutions because of its nonconvexity and discontinuity. Although the sparse coding has good performance, it may cause biased estimation since it penalizes true large coefficients more, and produce over-penalization. Consequently, it is necessary to find a
way to overcome the disadvantages of spare coding and sparse counting.

\begin{figure}[h]
\begin{center}
\begin{tabular}{c}
\includegraphics[width=0.48\textwidth,keepaspectratio]{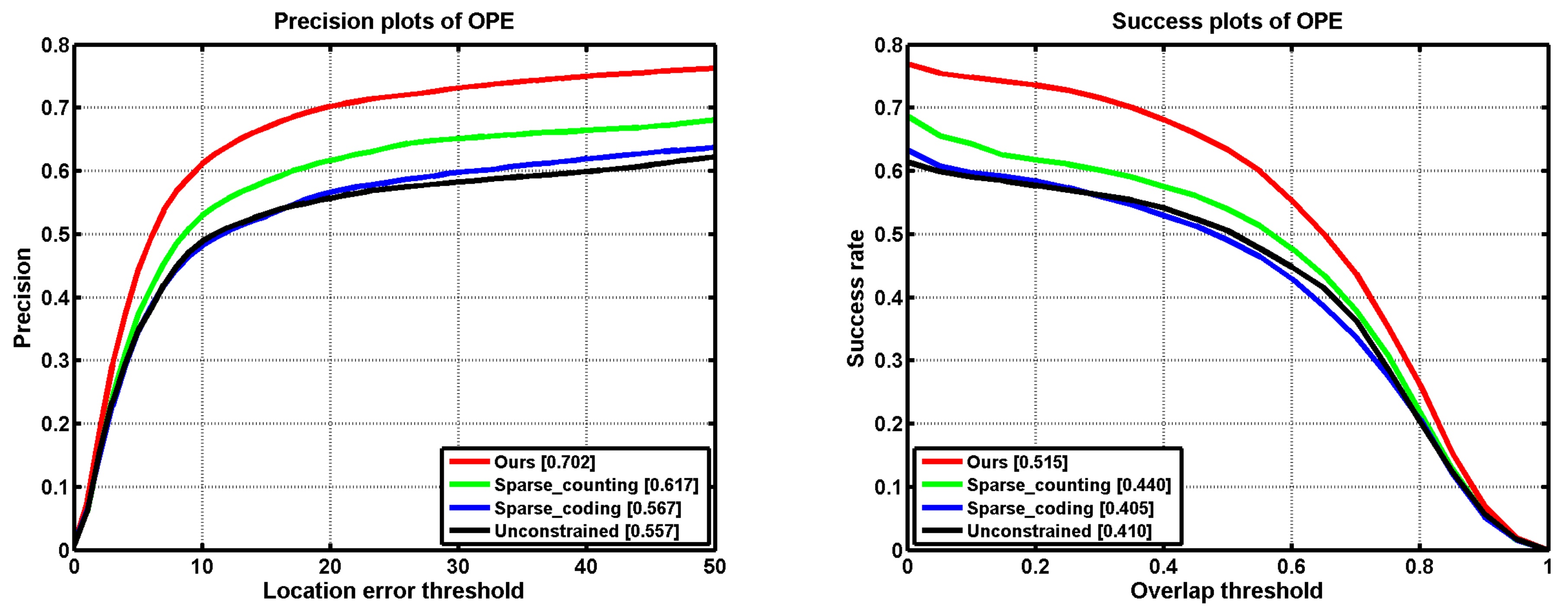}
\end{tabular}
\end{center}
\caption
{ \label{fig:precision_compare}
{\bf Precision and success plots of overall performance comparison among unconstrained regularization, sparse counting, sparse coding and ours for the 50 videos in the benchmark.} The mean precision scores are reported in the legends. }
\end{figure}

From the viewpoint of statistics, sparse representation are similar to variable selection when the dictionary is fixed. Besides, it is a bonus that Bayesian framework has been successfully applied to select variables by enforcing appropriate priors. Laplace priors were used to avoid overfitting and enforce sparsity in sparse linear model, which derives sparse coding problem. To further enforce sparsity and reduce over-penalization of sparse coding, each coefficient is assigned with a Bernoulli variable. Therefore, a novel model interpreted from a Bayesian perspective by carrying maximum a posteriori (MAP) is proposed, which turns out to be a combination of sparse coding and counting model. In paper \cite{Lu2013Sparse}, Lu \etal ~also consider $L_0$ and $L_1$ norm under a Bayesian perspective. However, considering that there will be occlusion, illumination change and background clutter in tracking, we restraint the noise with $L_1$ norm. Besides, We use an orthogonal dictionary to replace the redundant object templates as similar atoms of redundant templates may cause mistake of coefficients and huge computational complexity. Lastly, We propose closed solution of regularization which is the combination of the $L_0$ norm and $L_1$ norm. However Lu \etal ~obtain the approximate solution by using he Greedy Coordinate Descent.
%The new objective representation can generate more stable results than sparse counting and smaller reconstruction errors than the sparse coding.
%Xu et al. propose a tracking algorithm with structural local sparse model. But the template learnt in this method does not get ride of some drawbacks of subspace-based models, so the model is not effective for some sequences. From standpoint of computing complexity, we use compact dictionary instead templates. However, in orthogonal dictionary, there still has some background. Therefore, we propose a sparse model which combining the L0 norm and L1 norm to regularize the coefficients, which is able to remove the redundant parts.
\begin{figure*}[htb!]
\begin{center}
\begin{tabular}{c}
\includegraphics[width=0.98\textwidth,keepaspectratio]{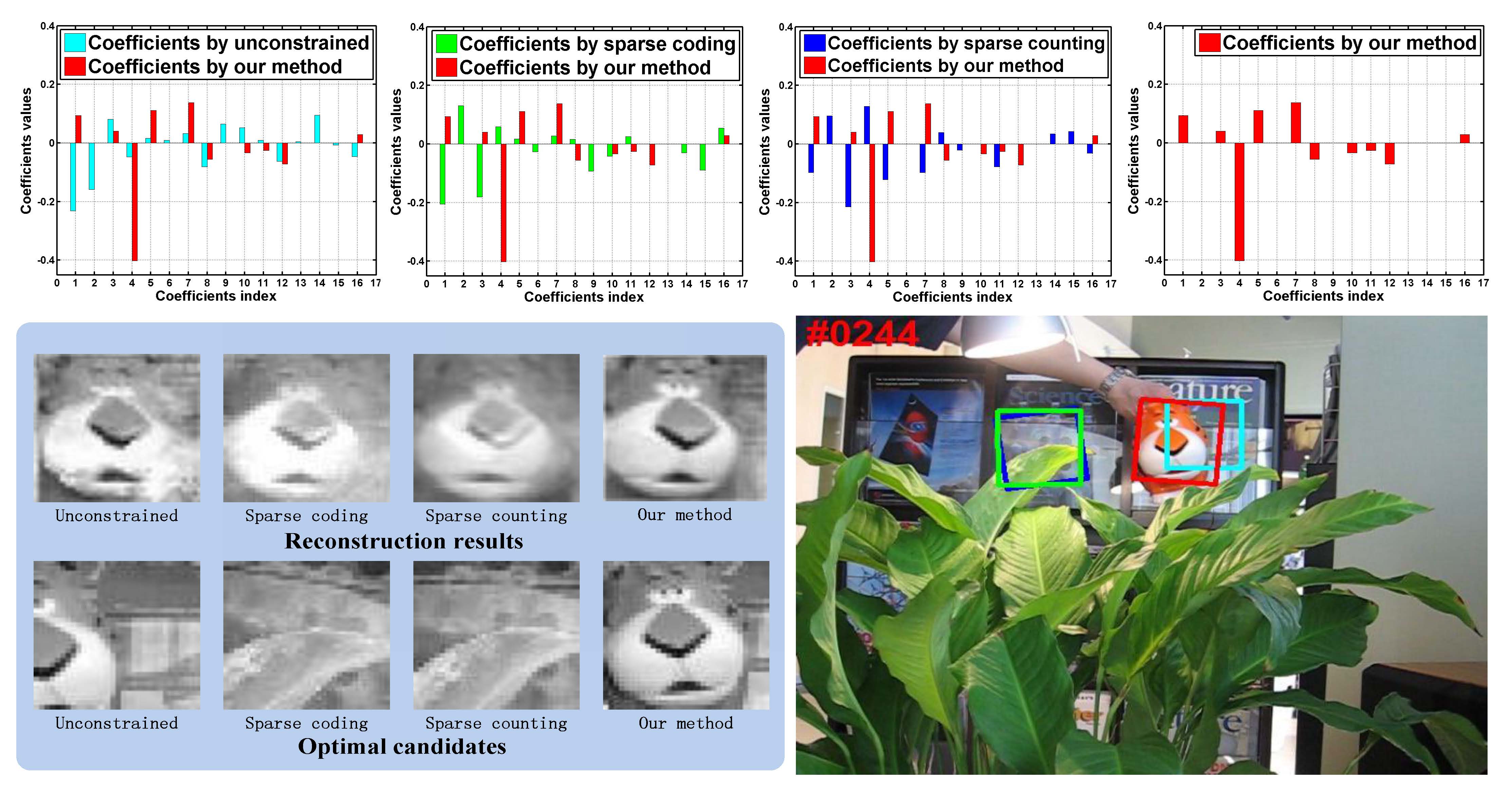}
\end{tabular}
\end{center}
\caption
{\label{fig:coefficient}
{\bf The comparison of coefficients, optimal candidates and reconstruction.} The top is the coefficients of our method versus unconstrained,
spars coding and sparse counting regularization, respectively. The bottom is the optimal candidates and reconstruction results by using unconstrained, sparse coding, sparse counting and our method under same dictionary, respectively. }
\end{figure*}
Tracking results by using unconstrained regularization, sparse counting, sparse coding and our model under the same dictionary $D$ are shown in Fig.~\ref{fig:coefficient}, respectively. As shown in Fig.~\ref{fig:coefficient}, one can see that the coefficients of unconstrained regularization and sparse coding are actually not sparse and the target object is not tracked well.
Similarly, sparse counting with sparsity coefficients sometimes cannot obtain appropriate linear combination of the orthogonal basis vectors, which will interfere with the tracking accuracy. However, we note that our method is able to reconstruct the object well and find the good candidate, then facilitating the tracking results. We also compare our model with unconstrained regularization, sparse counting, sparse coding over all 50 sequences in benchmark, the precision and success plots are shown in Fig.~\ref{fig:precision_compare}. One can see the parameter setting in the section Experimental Results.

{\flushleft \bf Contributions}: The contributions of this work are threefold.

(1) We propose a sparse coding and counting model from a novel Bayesian perspective for visual tracking. Compared to the state-of-the-art algorithms, the proposed method achieves more reliable tracking results.%$L_0$ and $L_1$ regularized representation of the target appearance

(2) We propose closed solution of combining the $L_0$ norm and $L_1$ norm based regularization in a unique one.

(3) Although the sparse coding and counting related minimization is an NP-hard problem,we show that the proposed model can be efficiently estimated by the proposed APG method.
This makes our tracking method computationally attractive in general and
comparable in speed with SP method~\cite{Lu/tip13/Wang13} and the accelerated $L_1$ tracker~\cite{APGL1/bao}.
%\begin{enumerate}[(1)]
%\item We propose a sparse coding and counting model from a novel Bayesian perspective for visual tracking. Compared to the state-of-the-art algorithms, the proposed method achieves more reliable tracking results.%$L_0$ and $L_1$ regularized representation of the target appearance
%\item We propose closed solution of combining the $L_0$ norm and $L_1$ norm based regularization in a unique one.
%
%\item Although the sparse coding and counting related minimization is an NP-hard problem,
%we show that the proposed model can be efficiently estimated by the proposed APG method.
%\end{enumerate}
\section*{Visual Tracking based on the Particle Filter}
\label{sec: Visual Tracking Algorithm based on the Particle Filter}

%Most visual tracking methods are based on the particle filter framework.
In this paper, we employ a particle filter to track the target object. The particle filter provides an estimate of posterior distribution
of random variables related to Markov chain. Given a set of observed image vectors $\by_t = \{\y_1, \y_2, ..., \y_t\}$ up to the $t$-th frame and target state
variable $\x_t$ that describes the six affine motion parameters, the posterior distribution $p(\mathbf{x}_{t}|\mathbf{Y}_{t})$  based on the Bayesian theorem is estimated by:
\begin{equation}
p(\mathbf{x}_{t}|\mathbf{Y}_{t})\propto p(\mathbf{y}_{t}|\mathbf{x}_{t})\int p(\mathbf{x}_{t}|\mathbf{x}_{t-1})p(\mathbf{x}_{t-1}|\mathbf{Y}_{t-1})d\mathbf{x}_{t-1},
\end{equation}
where $p(\y_t | \x_{t})$ is the observation model that estimates the likelihood of an observed image patch $\mathbf{y}_t$ belonging to the object class, and $p(\x_t | \x_{t-1})$ is the motion model that describes the state transition between consecutive frames.

\textbf{The Motion Model: }The motion model $p(\x_t | \x_{t-1}) = N (\x_t ; \x_{t-1}, \mathbf{\Sigma})$ models the parameters by independent Gaussian distribution around the counterpart in $\mathbf{x}_{t-1}$, where $\mathbf{\Sigma}$ is a diagonal covariance matrix whose elements are the variances of the affine parameters.
In the tracking framework, the optimal target state $\hat{\x}_t$ is obtained by the maximal approximate posterior (MAP) probability: $\hat{\x}_t = \arg\max_{\x_t^i}p(\x_t^i | \by_t )$,
where $\x_t^i$ indicates the $i$-th sample of the state $\x_t$.

\textbf{The observation model: }In this paper, we assume that the
tracked target object is generated by a subspace
(spanned by $\bd$ and centered at $\bmu$) with corruption (i.i.d Gaussian Laplacian noise),
$$\y = \bd\balpha+\mathbf{\epsilon}+\e,$$
where $\y\in \mathbb{R}^{N}$ denotes an observation vector centered at $\bmu$, the columns of $\bd=\{\d_1, \d_2,\ldots, \d_K\} \in \mathbb{R}^{N \times K}$ are orthogonal basis vectors of the subspace, $\balpha$ indicates the coefficients of basis vectors, $\mathbf{\epsilon}$ and $\e$ stand for the Gaussian noise and Laplacian noise vector respectively. the Gaussian component models small dense noise and the Laplacian one aims to handle outliers. As proposed by~\cite{Lu/CVPR13/Wang13}, under the i.i.d
Gaussian-Laplacian noise assumption, the distance between
the vector $\y$ and the subspace $(\bd, \bmu)$ is the least soft threshold squares distance:
$$d(\balpha, \e) = \min_{\balpha, \e}\frac{1}{2}\| \y-\bd\balpha - \e\|_2^2+ \lambda\|\e\|_1.$$
Thus, for each observation $\y_t$ corresponding to a predicted state $\x_t$, the observation model $p(\y_t | \x_{t})$ that is set to be
\begin{equation}\label{eq:observation-likelihood}
p(\y_t | \x_{t}) = \exp(-\tau d(\balpha^*, \e^*)),
\vspace{-0.66ex}
\end{equation}
where $\balpha^*$ and $\e^*$ are the optimal solution of Eq.~\eqref{eq:tracking modal} which will be introduced in detail
in next section, and $\tau$ is a constant controlling the shape of the Gaussian kernel.

\textbf{Model Update: }It is essential to update the observation model for handling appearance change of the target in visual tracking.
%It is noted that updating model is also very important in visual tracking.
%
Since the error term $\e$ can be used to identify some outliers (\eg, Laplacian noise, illumination),
we adopt the strategy proposed by~\cite{Lu/CVPR13/Wang13} to update the appearance model
using the incremental PCA with mean update~\cite{IVT/ijcv/RossLLY08} as follows,

\begin{equation}
\label{eq: model-update}
y_i = \left\{\begin{array}{ll}y_i, & \mbox{}\ e_i = 0,\\ \mu_i, & \mbox{}\
    $otherwise$, \end{array}\right.
\end{equation}
where $y_i$, $e_i$, and $\mu_i$ are the i-th elements of $\y$, $\e$, and $\bmu$, respectively,
$\bmu$ is the mean vector computed the same as~\cite{IVT/ijcv/RossLLY08}.

\section*{Object Representation under Bayesian Framework}

Based on the discussion in aforementioned Section, If $\y$ is viewed as the vectorized target region, it can be represented by an image subspace with corruption,
\begin{equation}\label{eq:1}
\y = \bd\balpha+\mathbf{\epsilon}+\e.
\end{equation}
~\cite{panl0} shows that sparse counting can remove redundant features (\eg, background portions) while selecting useful parts in the subspace. However, sparse counting will cause unstable solutions because of its nonconvexity and discontinuity. Sparse coding may produce over-penalization, despite its good stability.
%Consequently, it is necessary to find a way to overcome the disadvantages of the $L_0$ and $L_1$ regularization.
Considering that Bayesian framework has the capacity to encode prior knowledge and to make valid estimation of uncertainty, a novel model combining sparse coding and sparse counting is proposed for visual tracking. The model is
\begin{equation}\label{eq:tracking modal}
\min_{\balpha\,\e}\frac{1}{2}||\y-\bd\balpha-\e||_2^2+\beta||\e||_1+\lambda\gamma||\balpha||_1+\lambda(1-\gamma)||\balpha||_0,
\end{equation}
where $\bd^\top \bd = \mathbf{I}$, $\|\cdot\|_0$ denotes the $L_0$ norm which counts the number of non-zero elements,
$\|\cdot\|_2$ and $\|\cdot\|_1$ denote $L_2$ and $L_1$ norms, respectively,
$\gamma$, $\lambda$ and $\beta$ are regularization parameters, and $\mathbf{I}$ is an identity matrix.
The term $\|\e\|_1$ is used to reject outliers (\eg, occlusions), while $\|\balpha\|_0$ and $\|\balpha\|_1$ are used to select the useful subspace features.
%We note that if we set $\lambda = 0$,
%Eq.~\eqref{eq:tracking modal} is reduced to~\cite{Lu/tip13/Wang13}; if we set $\gamma = 0$, Eq.~\eqref{eq:tracking modal} is reduced to~\cite{panl0}.

Next we will introduce the aforementioned model under Bayesian framework in detail.
The joint posterior distribution of $\balpha, \r,\e$ and $\sigma^2$ based on the Bayesian theorem can be written as
%\end{equation}
\begin{align}\label{eq:PDF}
\begin{split}
&p(\balpha,\r,\e,\sigma^2|\bd,\y,\tilde{\mu},\tau_1,\tau_2,\kappa,\hat{\sigma})\propto \\ &p(\y|,\bd,\balpha,\r,\e,\sigma^2)p(\balpha|\sigma^2,\tilde{\mu})p(\r|\kappa)p(\e|\hat{\sigma})p(\sigma^2|\tau_1,\tau_2),\\
\end{split}
\end{align}
where $p(\y|\bd,\balpha,\r,\e,\sigma^2)$, $p(\balpha|\sigma^2,\tilde{\mu})$, $p(\r|\kappa)$, $p(\e|\hat{\sigma})$, $p(\sigma^2|\tau_1,\tau_2)$, denote the priors on the noisy vectorized target region, the coefficient vector $\balpha=[\alpha_{1},\alpha_{2},\ldots,\alpha_{K}]$, the index vector $\r = [r_{1}, r_{2},\ldots, r_{K}]$ ($r_{l} = \mathbb{I}(\alpha_{l} \neq 0),l = 1, 2,\ldots, K$), the Laplacian noise, and the noise level, respectively. In Eq.~\eqref{eq:PDF}, the parameters
$\tilde{\mu}$, $\tau_1$, $\tau_2$, $\hat{\sigma}$, and $\kappa$ are the relevant constant parameters of the priors.
%The priors are presented as follows.

With the definition of the index variable $r_{l}$ , Eq.~\eqref{eq:1} can be rewritten as
\begin{equation}
y_{j}=\sum_{l=1}^K d_{jl}r_{l}\alpha_{l}+\epsilon_{j}+e_{j},~j=1,2,\ldots,N.
\end{equation}
We generally assume that the noise $\epsilon_{j}$ follows the Gaussian distribution, $\ie,~p(\epsilon_{j})= N(0,\sigma^{2})$. We treat the Laplacian noise term $e_{j}$ as missing values with the same Laplacian prior. Therefore, the Prior $p(\y|,\bd,\balpha,\r,\e,\sigma^2)$ has the follow distribution:
\begin{align}
\begin{split}\label{eq:y}
&p(\y|,\bd,\balpha,\r,\e,\sigma^2)=\\
&\prod_{j=1}^NP(y_{j}|,\d_j,\balpha,\r,e_{j},\sigma^{2})\\
&=\prod_{j=1}^NN(\sum_{l=1}^K d_{jl}r_{l}\alpha_{l}+e_{j},\sigma^{2})\\.
\end{split}
\end{align}

To enforce sparsity, the coefficients $\balpha$ are assumed to follow Laplace distribution.
\begin{equation}\label{eq:balpha}
p(\balpha|\sigma^2,\tilde{\mu})=\prod_{l=1}^Kp(\alpha_{l}|\sigma^2,\tilde{\mu})= \prod_{l=1}^K\frac{1}{2\sigma^2\tilde{\mu}^{-1}}\exp(-\frac{|\alpha_{l}|}{\sigma^2\tilde{\mu}^{-1}}).
\end{equation}

Our goal is to remove redundant features while preserving the useful parts in the dictionary.
As Laplace priors resulting sparse coding may lead to over penalization on the large coefficients, we assume the index variable
$r_{l}$ of each coefficient $\alpha_{l}$ to be a Bernoulli variable to enforce sparsity and reduce over penalization.
\begin{equation}\label{eq:r}
p(\r|\kappa)=\prod_{l=1}^K\kappa^{r_{l}}(1-\kappa)^{1-r_{l}},
\end{equation}
where $\kappa \leq 1/2$. Here, the Bernoulli prior on $r_{l}$ means that $r_{l}$ will have probability $\kappa$ to be 1 and $1-\kappa$ to be 0, if the prior information is known.

The noise $e_{j}$ is aims at handling outliers, so it follows Laplace distribution:
\begin{equation}\label{eq:e}
p(\e|\hat{\sigma})=\prod_{j=1}^{N}p(e_{j}|\hat{\sigma})= \prod_{j=1}^{N}\frac{1}{2\hat{\sigma}}\exp(-\frac{|e_{j}|}{\hat{\sigma}}).
\end{equation}

The variances of noises are assigned with Inverse Gamma prior as follow:
\begin{equation}\label{eq:sigma}
p(\sigma^2|\tau_1,\tau_2)=\frac{\tau_2^{\tau_1}}{\Gamma(\tau_1)}\sigma^{-2(\tau_1+1)}\exp(-\frac{\tau_2}{\sigma^2}),
\end{equation}
where $\Gamma(\cdot)$ denotes the gamma function.

Then, the optimal $\balpha, \r,\e,\sigma^2$ are obtained by the MAP probability. After taking the negative logarithm, the formula is
%\begin{equation}
%(\balpha^*,\r^*,\e^*,\sigma^{*2})={\arg\min}_{\balpha,\r,\e,\sigma^2}\{-2logp(\balpha,\r,\e,\sigma^2|\bd,\y,\tilde{\mu},\tau_1,\tau_2,\kappa,\hat{\sigma})\}.
%\end{equation}
\begin{equation}
\begin{aligned}
&(\balpha^*,\r^*,\e^*,\sigma^{*2})=\arg\min\limits_{\balpha,\r,\e,\sigma^2}   \\
& \{-2logp(\balpha,\r,\e,\sigma^2|\bd,\y,\tilde{\mu},\tau_1,\tau_2,\kappa,\hat{\sigma})\}.
\end{aligned}
\end{equation}

Combining the aforementioned Eq.~\eqref{eq:PDF}, Eq.~\eqref{eq:y}, Eq.~\eqref{eq:balpha}, Eq.~\eqref{eq:r}, Eq.~\eqref{eq:e}, Eq.~\eqref{eq:sigma}, we have
%\begin{align}\label{eq:2}
%\begin{split}
%&-2logp(\balpha,\r,\e,\sigma^2|\bd,\y,\tilde{\mu},\tau_1,\tau_2,\kappa,\hat{\sigma})=\\
%&\frac{1}{\sigma^2}\sum_{j=1}^N(y_{j}-\sum_{l=1}^{K}d_{jl}r_{l}\alpha_{l}-e_{j})^2+\frac{1}{\sigma^2}\frac{2\sigma^2}{\hat{\sigma}}\sum_{j=1}^{N}|e_{j}|+\frac{2\tilde{\mu}}{\sigma^2}\sum_{l=1}^K|\alpha_{l}|+\\
%&(2N+2K+2\tau_1+2)log\sigma^2+\frac{2\tau_2}{\sigma^2}+\sum_{l=1}^{K}r_{l}log\frac{(1-\kappa)^2}{\kappa^2}+const.\\
%\end{split}
%\end{align}
\begin{align}\label{eq:2}
\begin{split}
&-2logp(\balpha,\r,\e,\sigma^2|\bd,\y,\tilde{\mu},\tau_1,\tau_2,\kappa,\hat{\sigma})=\\
&\frac{1}{\sigma^2}\sum_{j=1}^N(y_{j}-\sum_{l=1}^{K}d_{jl}r_{l}\alpha_{l}-e_{j})^2+\frac{1}{\sigma^2}\frac{2\sigma^2}{\hat{\sigma}}\sum_{j=1}^{N}|e_{j}|\\
&+\frac{2\tilde{\mu}}{\sigma^2}\sum_{l=1}^K|\alpha_{l}|+(2N+2K+2\tau_1+2)log\sigma^2+\frac{2\tau_2}{\sigma^2}\\
&+\sum_{l=1}^{K}r_{l}log\frac{(1-\kappa)^2}{\kappa^2}+const.\\
\end{split}
\end{align}
With fixing $\sigma^2 = 1$, Eq.~\eqref{eq:2} can be rewritten as
\begin{equation}\label{eq:3}
||\y-\bd \balpha-\e||_2^2+2\beta||\e||_1+2\tilde{\mu}||\balpha||_1+\rho_{\kappa}||\balpha||_0+const,
\end{equation}
where $\rho_{\kappa}= log (1 - \kappa)^2/\kappa^2, \beta=\sigma^2/\hat{\sigma}$. With $\gamma \in [0, 1], \lambda = \tilde{\mu} + 1/2\rho_{\kappa}$ and $\gamma = 4\tilde{\mu}/(2\tilde{\mu} + \rho_{\kappa})$, Eq. \eqref{eq:3} can be rewritten as
\begin{equation}\label{eq:4}
\frac{1}{2}||\y-\bd\balpha-\e||_2^2+\beta||\e||_1+\lambda\gamma||\balpha||_1+\lambda(1-\gamma)||\balpha||_0+const,
\end{equation}
By observing the objective function in Eq.~\eqref{eq:4}, it can be found that the essential regularization in Eq.~\eqref{eq:4} is a combination of the sparse coding and the sparse counting.
With a fixed appropriate orthogonal dictionary D, Eq.~\eqref{eq:4} can be written as an optimization problems Eq.~\eqref{eq:tracking modal}.
%\begin{equation}\label{eq:5}
%\begin{aligned}
%&\min_{\balpha,\e}\frac{1}{2}||\y-\bd\balpha-\e||_2^2+\beta||\e||_1+\lambda\gamma||\balpha||_1+\lambda(1-\gamma)||\balpha||_0,\\
%& s.t.~~ \hat{\bd}^\top\hat{\bd}=\mathbf{\I}.
%\end{aligned}
%\end{equation}

%----------------------------------------------------------------------------------
\section*{Theory of Fast Numerical Algorithm}

As we know, APG is an excellent algorithm for convex programming~\cite{APG/lin,APG/Tseng} and has been used in visual tracking. In this section, we propose a fast numerical algorithm for solving the proposed nonconvex and nonsmooth model by using APG approach. The experimental results show that it can converge to a solution quickly and achieve attractive performance.
%Recently, there already exist some research ensuring APG can converge to a critical point in nonconvex programming by introducing a monitor~\cite{li2015accelerated}.
%However, APG approach we proposed can also ensure convergence without introducing something else in nonconvex programming. Considering the research is still in process, so we only present the conclusion.
Besides, the closed solution of the combining $L_0$ and $L_1$ based regularization is provided.
%We also show that the closed solution has great sparsity and efficiency.

\subsection*{APG Algorithm for Solving Eq.~\eqref{eq:tracking sub-modal}}
Eq.~\eqref{eq:tracking modal} contains two subproblem: one is solving $\balpha$ given fixed $\e$, the other one is solving $\e$ given fixed $\balpha$, the formula is shown as follow
\begin{equation}\label{eq:tracking sub-modal}
\left\{\begin{array}{l}
\balpha=\arg\min\limits_{\balpha}\frac{1}{2}||\y-\bd\balpha-\e||_2^2+\lambda\gamma||\balpha||_1+\lambda(1-\gamma)||\balpha||_0,\\
\e=\arg\min\limits_{\e}\frac{1}{2}||\y-\bd\balpha-\e||_2^2+\beta||\e||_1.
\end{array}\right.
\end{equation}

Solving Eq.~\eqref{eq:tracking sub-modal} is an NP-hard problem because it involves a discrete counting metric.
%However, the orthogonality of $D$ makes solving~\eqref{eq:l0-model} tractable (see analysis in Sec.~\ref{ssec: Analysis-on-the-Effectiveness}).
We adopt a special optimization strategy based on the APG approach~\cite{APG/lin}, which ensures each step be solved easily. In APG Algorithm, we need to solve
\begin{equation}\label{eq:apg-model-minimization-1}
\left\{\begin{array}{l}
\balpha_{k+1}^*=\arg\min\limits_{\balpha} \lambda\gamma\|\balpha\|_1+\lambda(1-\gamma) \|\balpha\|_0+\frac{L}{2} \| \balpha  - \z_{k+1}^{\balpha} + \frac{\nabla_\balpha F(\z_{k+1})}{L}\|_2^2,\\
\e_{k+1}^*= \arg\min\limits_{\e}\beta\| \e \|_1 + \frac{L}{2}\| \e  - \z_{k+1}^{\e} + \frac{\nabla_\e F(\z_{k+1})}{L}\|_2^2,
\end{array}\right.
\end{equation}
where $\z_{k+1}=(\z_{k+1}^{\balpha},\z_{k+1}^{\e})$, $\nabla_{\balpha} F(\balpha, \e) = \bd^\top(\bd\balpha + \e - \y)$,
$\nabla_\e F(\balpha, \e) = \e - (\y - \bd\balpha)$, and $L$ is a Lipschitz constant.

The solutions of Eq.~\eqref{eq:apg-model-minimization-1} can be obtained by
\begin{equation}\label{eq:apg-model-minimization-1-sol}
\left\{\begin{array}{l}
\balpha_{k+1}^*= \mathcal{E}_{(\lambda\gamma/L,\lambda(1-\gamma)/L)}\left( \z_{k+1}^{\balpha} - \frac{\nabla_\balpha F(\z_{k+1})}{L}\right),\\
\e_{k+1}^{*}= \mathcal{S}_{\beta/L}\left( \z_{k+1}^{\e} - \frac{\nabla_\e F(\z_{k+1})}{L}\right),
\end{array}\right.
\end{equation}
where $\mathcal{S}_{\theta}(y) = \text{sign}(y)\max(|y| - \theta,0)$,
and $\mathcal{E}_{(\delta,\eta)}(y)$ is defined as
\begin{equation}
\mathcal{E}_{(\delta,\eta)}(y)=\left\{
\begin{array}{ll}
y-\delta, & \mbox{}\ y > \delta+\sqrt{2\eta},\\
y+\delta, & \mbox{}\ y < -\delta-\sqrt{2\eta},\\
0,& \mbox otherwise.
\end{array}\right.
\end{equation}

The numerical algorithm for solving Eq.~\eqref{eq:tracking sub-modal} is summarized in Algorithm~\ref{alg:our-apg-alogrithm}. Due to the orthogonality of $\bd$, Algorithm~\ref{alg:our-apg-alogrithm} converges fast, and its computation cost does not increase compared to the solver of $L_1$ regularized model.
\begin{algorithm}
\caption{Fast numerical algorithm for solving Eq.~\eqref{eq:tracking sub-modal}}
\label{alg:our-apg-alogrithm}
\begin{algorithmic}
\STATE \textbf{Initialize:} Set initial guesses $\balpha_0 = \balpha_{-1}  = \textbf{0}$, $\e_0 = \e_{-1} = \textbf{0}$, and $t_0 = t_{-1} = 1$.
\STATE \textbf{while} not convergence or termination \textbf{do}
\STATE \textbf{Step 1:} $\z_{k+1}^{\balpha}:= \balpha_k + \frac{t_{k-1} -1 }{t_{k}}(\balpha_k - \balpha_{k-1})$;
\STATE \textbf{Step 2:} $\z_{k+1}^{\e}:= \e_k + \frac{t_{k-1} -1 }{t_{k}}(\e_k - \e_{k-1})$;
\STATE \textbf{Step 3:} $\balpha_{k+1}= \mathcal{E}_{(\lambda\gamma/L,\lambda(1-\gamma)/L)}\left( \z_{k+1}^{\balpha} - \frac{\nabla_\balpha F(\z_{k+1})}{L}\right)$;
\STATE \textbf{Step 4:} $\e_{k+1}= \mathcal{S}_{\beta/L}\left( \z_{k+1}^{\e} - \frac{\nabla_\e F(\z_{k+1})}{L}\right)$;
\STATE \textbf{Step 5:} $t_{k+1}:= \frac{1+\sqrt{1+4t_k^2}}{2}$, $k\leftarrow k+1$.
\STATE \textbf{end while}\\
%\CHECK{[JWL: only $\nabla$ is used in the equations instead of $\nabla_{\delta}$ or $\nabla_e$]}
%Js: Ok
\end{algorithmic}
\end{algorithm}

\subsection*{Closed Solution of combining $L_1$ and $L_0$ regularization}
This subsection mainly focus on a sparse combinatory model which combines $L_0$ and $L_1$ norm together as the regularizer term
\begin{equation}\label{eq:problem}
\min_x \frac{1}{2}(x-y)^2+\delta|x|+\eta|x|_0,
\end{equation}
where $x,y\in\mathbb{R}^1$, and $|x|$ denotes $L_0$ norm: if $x = 0$, then$|x|_0=0$, and $|x|_0=1$, otherwise.

\vspace{2ex}\noindent{\footnotesize\textbf{lemma.}
\label{lemma:1}
The optimal solution $x^*$ of the Eq.~\eqref{eq:problem} is defined as
\begin{equation}
\label{eq: proof}
x^* = \left\{
\begin{array}{ll}
y-\delta, & \mbox{}\ y > \delta+\sqrt{2\eta},\\
y+\delta, & \mbox{}\ y < -\delta-\sqrt{2\eta},\\
0,& \mbox otherwise.
\end{array}\right.
\end{equation}}

The proof can be found in Supporting Information. If $x\in\mathbb{R}^N$, the Eq.~\eqref{eq:problem} changes into
\begin{equation}\label{eq:regression}
\min_\x \frac{1}{2}||\x-\y||_2^2+\delta||\x||_1+\eta||\x||_0,
\end{equation}
where $||\x||_1=\sum_{i=1}^N|x_i|$ and $||\x||_0=\sum_{i=1}^N|x_i|_0$. It is obvious that Eq.~\eqref{eq:problem} can be turned into
\begin{equation}
\min_{x_i} \sum_{i=1}^N \frac{1}{2}(x_i-y_i)^2+\delta|x_i|+\eta|x_i|_0.
\end{equation}
So it can be seen as a sequence of optimization of $x_i,i=1,\ldots,n$, and each can be solved by Lemma.
More analysis about combination of $L_1$ and $L_0$ regularization can be found in Supporting Information.
\subsection{Analysis of the combinatory model Eq.~\eqref{eq:regression}}

\begin{figure}[h]
\begin{center}
\begin{tabular}{c}
\includegraphics[width=0.49\textwidth,keepaspectratio]{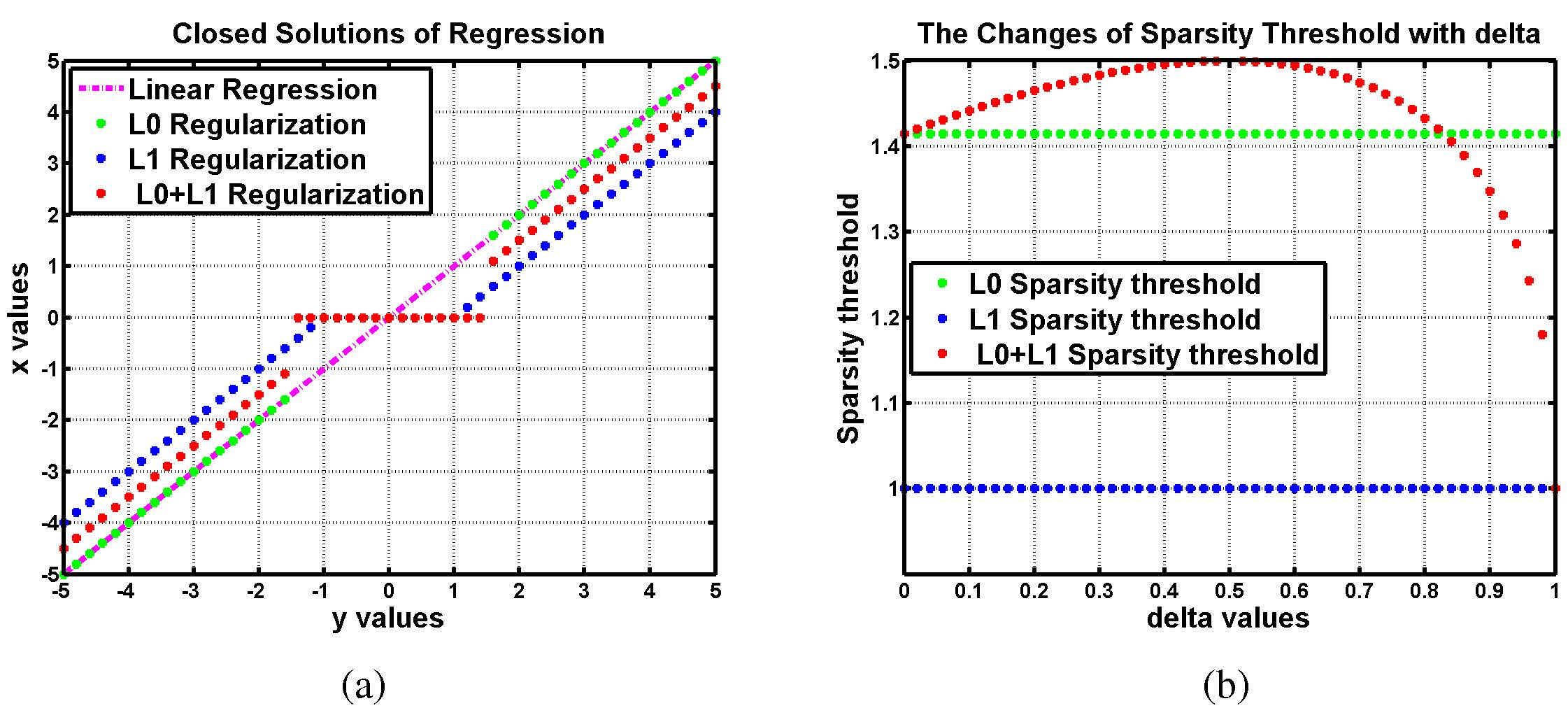}\label{fig:Regularized Regression}
\end{tabular}
\end{center}
\caption{{\bf Analysis about combination of $L_1$ and $L_0$ regularization.} (a) shows the closed solutions of linear regression, $L_0$, $L_1$, $L_0+L_1$ regularized regression, respectively. (b) shows the sparsity threshold changes of $L_0$, $L_1$ and $L_0+L_1$ regularized regression, respectively.}\label{fig: Regression}
\end{figure}
In Eq.~\eqref{eq:regression}, if we set $\delta=0$ and $\eta=0$, the model degenerates to the linear regression. If we set $\delta=0$, Eq.~\eqref{eq:regression} reduces to $L_0$ regularized regression, while becoming $L_1$ regularized regression when $\eta=0$. S2 Fig.~\ref{fig: Regression} (a) shows the closed solutions of these four cases. We set $\delta=\eta=0.5$ in Eq.~\eqref{eq:regression} ($L_0+L_1$ regularized regression), $\eta=1$ in $L_0$ regularized regression, and $\delta=1$ in $L_1$ regularized regression. We note that $L_0+L_1$ regularized regression has the same sparsity as $L_0$ regularized regression, while causing little over penalization than $L_1$ regularized regression. In S2 Fig.~\ref{fig: Regression} (b), sparsity threshold changes of $L_0$, $L_1$ and $L_0+L_1$ regularized regression are shown, respectively. When $\delta = 1-\eta$ changes from 0 to 1, the sparsity threshold of $L_0+L_1$ varies from that of $L_0$ to the threshold of $L_1$. Besides, it is obvious that the threshold of $L_0+L_1$ is larger than those of $L_0$ and $L_1$ in interval $(0, 0.8]$.
%
%\subsection{Analysis of the combinatory model}
%In Eq.~\eqref{eq:regression}, if we set $\delta=0$ and $\eta=0$, the model degenerates to the linear regression. If we set $\delta=0$, Eq.~\eqref{eq:regression} reduces to $L_0$ regularized regression, while becoming $L_1$ regularized regression when $\eta=0$. Fig.~\ref{fig: Regression}(a) shows the closed solutions of these four cases. We set $\delta=\eta=0.5$ in Eq.~\eqref{eq:regression} ($L_0+L_1$ regularized regression), $\eta=1$ in $L_0$ regularized regression, and $\delta=1$ in $L_1$ regularized regression. We note that $L_0+L_1$ regularized regression has the same sparsity as $L_0$ regularized regression, while causing little over penalization than $L_1$ regularized regression. In Fig.~\ref{fig: Regression}(b), sparsity threshold changes of $L_0$, $L_1$ and $L_0+L_1$ regularized regression are shown, respectively. When $\delta = 1-\eta$ changes from 0 to 1, The sparsity threshold of $L_0+L_1$ varies from that of $L_0$ to the threshold of $L_1$. Besides, it is obvious that the threshold of $L_0+L_1$ is larger than those of $L_0$ and $L_1$ in interval $(0, 0.8]$.
%
\section*{Orthogonal Dictionary learning for Visual Tracking}
In this section, we demonstrate dictionary learning in detail through three parts: dictioanry initialization, orthogonal dictionary update and dictionary reinitialization.

\begin{figure}[htbp]
\begin{center}
\begin{tabular}{c}
\includegraphics[width=0.49\textwidth]{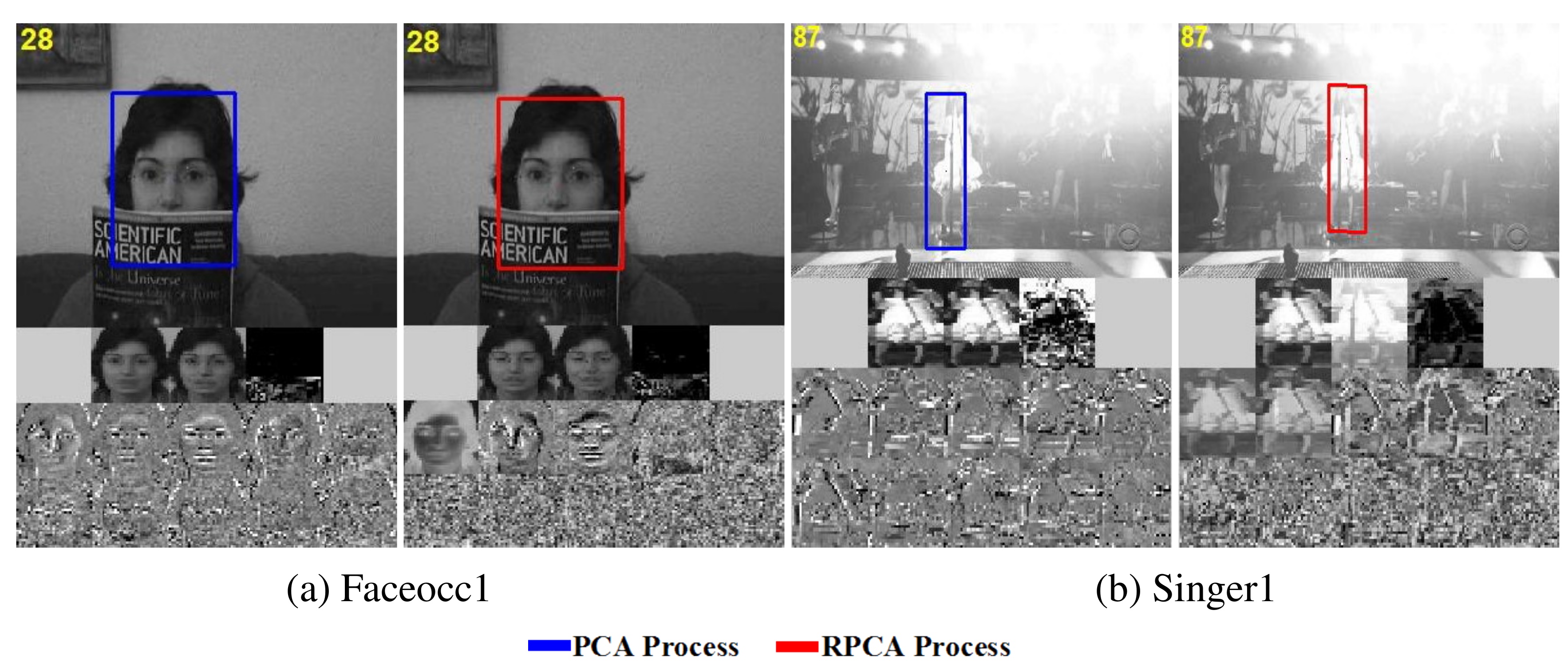}
\end{tabular}
\end{center}
\caption{{\bf Comparison of PCA process to RPCA process.} The upper portion of the image is the tracking frame. the middle of the image consists of three sub-pictures, the left is the mean image, the middle is the reconstruction result, and the right is the Lapalace noise. the bottom of the image is the top ten basis vectors of dictionary. (a) shows the tracking results of PCA and RPCA dictionary initialization. The tracking performance with and without RPCA reinitialization is shown in (b). }
%Fig.~\ref{fig:process/PCA_VS_RPCA}(a) shows the PCA process and RPCA process in initialization of the dictionary basis. When the initial frames have no Lapalace noise, there is no need to doing RPCA process as both of them can separate the noise (\eg, occlusion) from the optimal candidates and obtain great reconstruction results. In Fig.~\ref{fig:process/PCA_VS_RPCA} (b), the object undergoes variable illumination, our tracker drifts away and the noise becomes not sparse with PCA process. We use RPCA which can obtain the intrinsic low rank features of the subspace to reinitialize dictionary during the tracking process and retrack the object.
\label{fig:process/PCA_VS_RPCA}
\end{figure}

\textbf{Dictioanry Initialization:}\label{Dictioanry Initialization}
There are two schemes to initialize the orthogonal dictionary, one is doing PCA for the set of initial first $k$ frames $\by_k$, the other is doing RPCA for $\by_k$. When initial frames do not undergo corruption (\eg,~occlusion or illumination), we do PCA for $\by_k$ instead of RPCA. The whole process of PCA is doing skinny SVD for $\by_k$ and get the basis vectors of column space as the initial dictionary. However, when initial frames have large sparse noise, RPCA is selected to get the intrinsic low-rank features $\mathbf{Z}_k$, which can be obtained by solving~\cite{LRF/Zhang14}:
\begin{equation}
\min\limits_{\mathbf{Z}_k,\mathbf{E}_k} \|\mathbf{Z}_k\|_* + \lambda\|\mathbf{E}_k\|_1, \ s.t. \ \mathbf{Y}_k=\mathbf{Z}_k + \mathbf{E}_k.\label{eq:rpca_ini}
\end{equation}
When solving Eq.~\eqref{eq:rpca_ini}, the skinny SVD of $\mathbf{Z}_k$ is readily available: $\mathbf{Z}_k=\mathbf{U}_k\Sigma_k\mathbf{V}_k^T$, and $\bd=\mathbf{U}_k$ is the initial orthogonal dictionary.
%As the analysis in \cite{L1PCA/Liu14}, the skinny SVD of $\mathbf{Z}_k$ is readily available when solving Eq.~\eqref{eq:rpca_ini}:
Fig. \ref{fig:process/PCA_VS_RPCA} (a) shows that PCA initialization and RPCA initialization both perform well when the initial first $k$ frames have little noise. The initial frames is generally clean, therefore, we choose PCA initialization as the default.
\textbf{Orthogonal Dictionary Update:}\label{dictionary update}
As the appearance of a target may change drastically, it is necessary to update the orthogonal dictionary $\bd$. Here we adopt an incremental PCA algorithm \cite{Levey2000SKL} to update the dictionary.
%Suppose we have calculated the dictionary $\bd$ for $\mathbf{Y}_t$, then we aim at updating it for the concatenation $\mathbf{Y}_{t+m}=[\mathbf{Y}_t,\mathbf{Y}_{t+1:t+m}]$, where $\mathbf{Y}_{t+1:t+m}$ are the optimal candicates with mean update of the next $m$ frames.

\textbf{ Dictionary reinitialization:}\label{Reinitializing Dictionary}
When the tracker is prone to drift, dynamically reinitializing dictionary to obtain the intrinsic subspace features is needed. We adopt the strategy proposed by~\cite{LRF/Zhang14}. The reinitialization is performed at $t$-th frame if $\sigma=\|\e_t\|_0/len(\e_t) > thr$, where $\e_t$ is the noise item at $t$-th frame, $len(.)$ is the length of vector, and $thr>0$ is a threshold parameter (generally 0.5). If $\sigma > thr $, we reinitialize the dictionary in the same way as initialization of dictionary by doing RPCA, but $\mathbf{Y}_t$ in Eq.~\eqref{eq:rpca_ini} is different. Here, $\mathbf{Y}_t$ consists of optimal candidate observations respectively
from the initial $n$ (generally 10) frames and the latest $t-n$ frames (we set $t=30$).
Fig.~\ref{fig:process/PCA_VS_RPCA} (b) compares the tracking performance within and without RPCA reinitialization when the object undergoes variable illumination. After reinitializing dictionary, our tracker retracks the object, so reinitializing dictionary is efficient to improve the reconstruction ability. In Algorithm~\ref{alg:tracker}, we summarize the overall tracking process for frame $t$.

\section*{Experimental Results}
In this section, we compare the performance of our proposed tracker with several state-of-the-art tracking algorithms, such as TLD~\cite{TLD/pami/KalalMM12}, IVT~\cite{IVT/ijcv/RossLLY08}, ASLA~\cite{ALSA/cvpr/Lu/JiaLY12}, $L_1$APG~\cite{APGL1/bao}, MTT~\cite{MTT/ijcv/ZhangGLA13}, SP~\cite{Lu/tip13/Wang13}, SPOT~\cite{zhang2013structure}, FOT~\cite{vojivr2014enhanced}, SST~\cite{zhang2015structural}, SCM~\cite{SCM/cvpr/Lu/ZhongLY12}, MIL~\cite{MIL/cvpr/BabenkoYB09}, and Struck~\cite{Struck/iccv/HareST11}, on a benchmark~\cite{survey2/cvpr/Yang} with 50 challenge video sequences. Our tracker is implemented in MATLAB and runs at 4.2 fps on an Intel 2.53 GHz Dual-Core CPU with 8GB memory, running Windows 7 and Matlab (R2013b). We empirically set $\eta = 0.1$, $\lambda  = 0.5$, $\gamma = 0.1$, $\tau = 0.05$  and the Lipschitz constant L = 2. Before solving Eq.~\eqref{eq:tracking modal}, all the candidates $\y$ are centralized.
Considering the efficiency, the updated orthogonal dictionary $\bd$ is taken columns corresponding to the $16$ largest eigenvalues of PCA or RPCA, 600 particles are adopted, and the model is incrementally updated every $5$ frames.
In the following, we present both qualitative and quantitative comparisons of above mentioned methods.
\begin{algorithm}[hb]
   \caption{Robust Visual Tracking Using Our tracker}
   \label{alg:tracker}
\begin{algorithmic}
    \STATE {\bfseries Initialization:} Initialize orthogonal dictionary $\bd$ by performing PCA on ${{\bf{Y}}_{k}}$.
    \STATE {\bfseries Input:} State $\mathbf{x}_{t-1}$ ($t>k$) and orthogonal dictionary $\bd$.
    \STATE {\bfseries Step 1:} Draw new samples $\mathbf{x}_t^i$ from $\mathbf{x}_{t-1}$ and
  obtain corresponding candidates $\mathbf{y}_t^i$.
    \STATE {\bfseries Step 2:} Obtain $\mathbf{\alpha}_{t}^i$ and $\mathbf{e}_{t}^{i}$ using (\ref{eq:tracking sub-modal}).
    \STATE {\bfseries Step 3:} For each candidate, calculate the observation probability $p(\mathbf{y}_t^i|\mathbf{x}_{t}^i)$ using (\ref{eq:observation-likelihood}).
    \STATE {\bfseries Step 4:} Find the tracking result patch $\mathbf{y}_t^*$ with the maximal observation likelihood and its corresponding noise $\mathbf{e}_{t}^*$.
    \STATE {\bfseries Step 5:} perform an
    incremental PCA algorithm to update the orthogonal dictionary $\bd$ every five frames. If $\sigma>thr$, reinitializing Dictionary at $t$-th frame using \eqref{eq:rpca_ini}.
    \STATE {\bfseries Output:} State $\mathbf{x}_t^*$ and corresponding image patch; orthogonal dictionary $\bd$.
\end{algorithmic}
\end{algorithm}
%\subsection*{Ethics Statements}
%The images of all Figures are from public tracking database "Visual Tracker Benchmark"  and free from copyright. The URL of racking database is \url{http://www.visual-tracking.net}. This website contains data and code of the benchmark evaluation of online visual tracking algorithms. You can find the following resources from this site: Benchmark results, Dataset with ground-truth annotations, Code library. The reference of the database can refer to Wu \etal~\cite{survey2/cvpr/Yang} in detail.
 %Therefore, it is not necessary to include required ethics statements.

%More results can also be found in supplemental materials (videos in avi format).
%\subsection*{Ethics Statement}
%
%The study was approved by the authors' Institutional Review Board (IRB), and has been conducted according to the principles expressed in the?Declaration of Helsinki.

\subsection*{Qualitative Evaluation}
\begin{figure*}[htb]
\begin{center}
\begin{tabular}{c}
\includegraphics[width=0.99\textwidth,keepaspectratio]{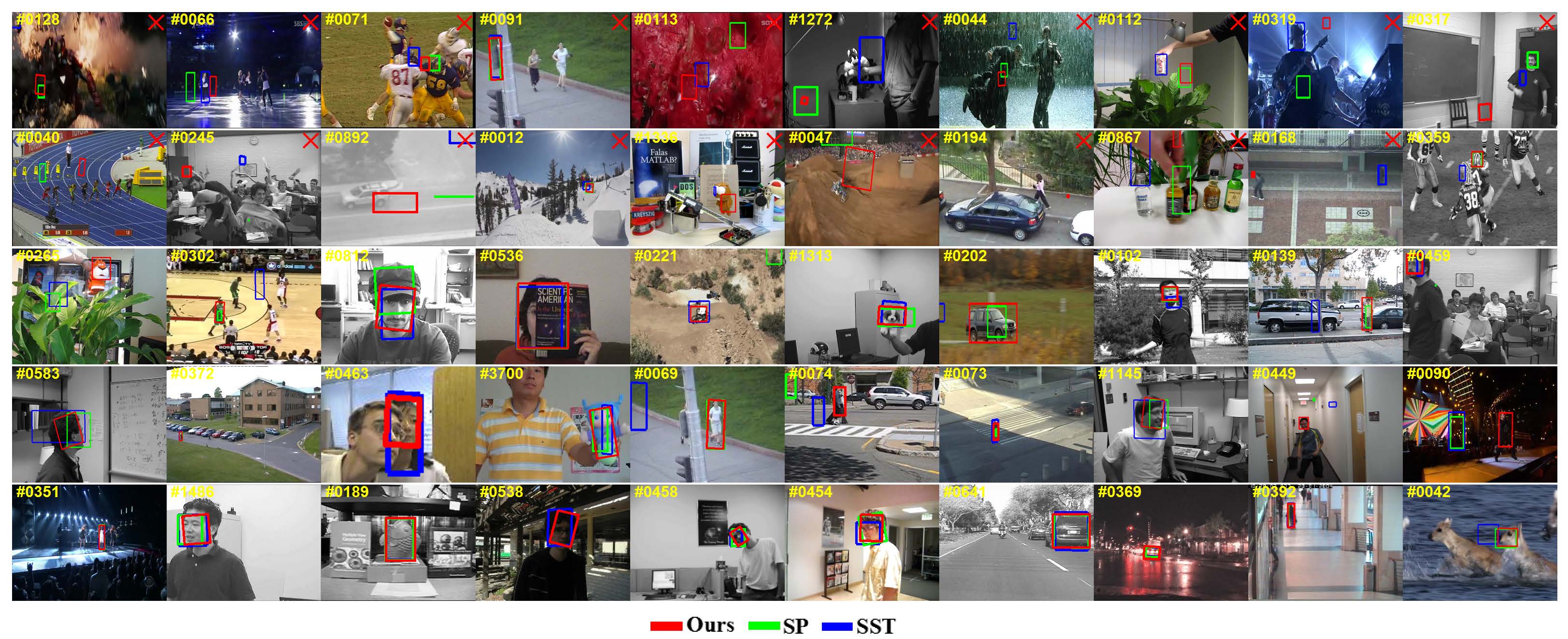}
\end{tabular}
\end{center}
\caption{\label{fig:all}{\bf Qualitative results for our method, compared with SP and SST.} Reprinted from~\cite{survey2/cvpr/Yang} under a CC BY license, with permission from Yi Wu, original copyright 2013.}
\end{figure*}

\begin{figure*}[htb!]
\begin{center}
\begin{tabular}{c}
\includegraphics[width=0.99\textwidth,keepaspectratio]{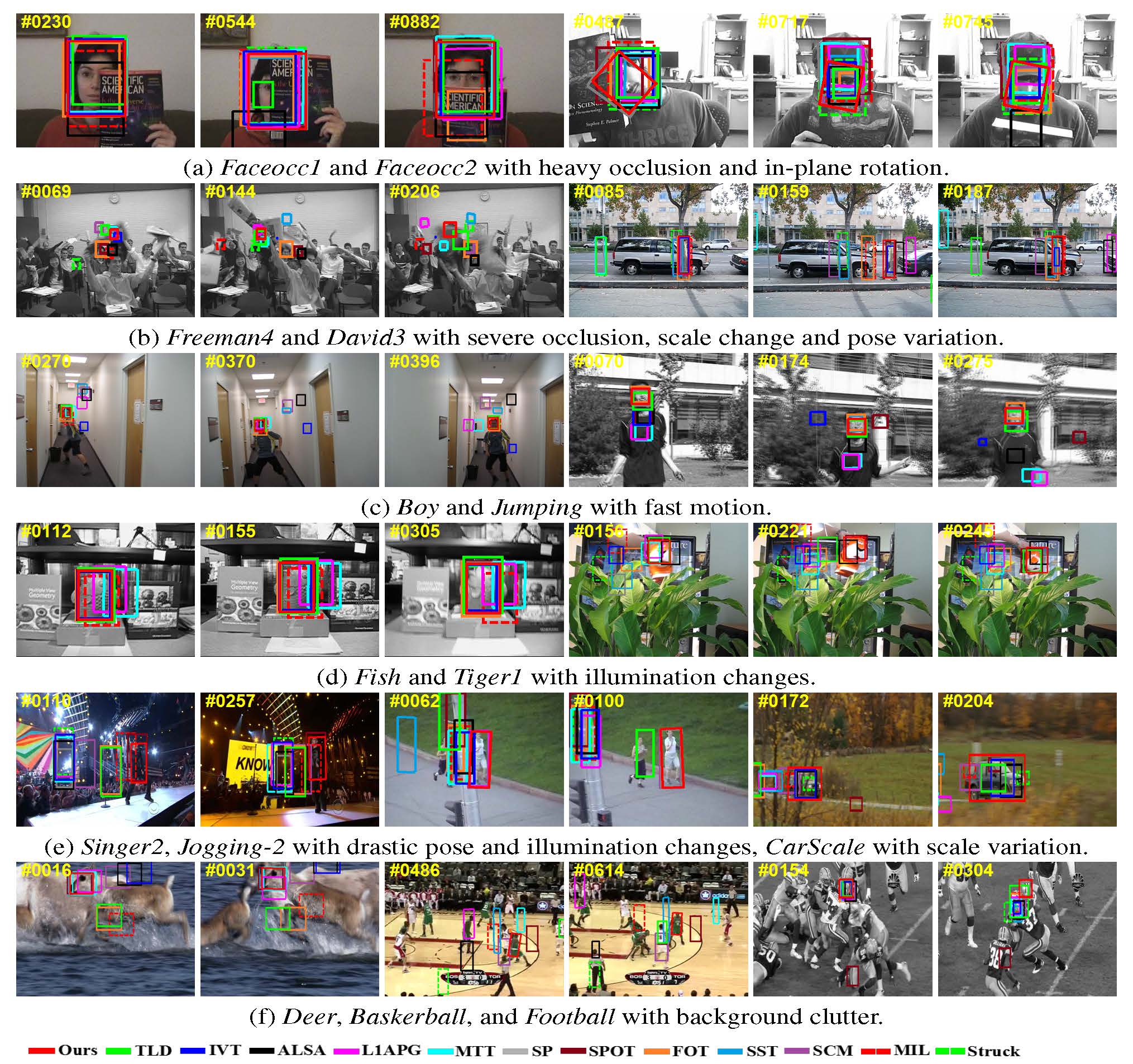}
\end{tabular}
\end{center}
\caption{\label{fig:trackingResult}{\bf Sampled tracking results of evaluated algorithms on fourteen challenging image sequences.} Reprinted from~\cite{survey2/cvpr/Yang} under a CC BY license, with permission from Yi Wu, original copyright 2013.}
\end{figure*}

Fig.~\ref{fig:all} were taken the frames of the 50 videos to show the Qualitative results for our method, compared with the top-performing SP and SST. We choose some examples from part of 50 sequences to illustrate the effectiveness of our method. Fig.~\ref{fig:trackingResult} shows the visualization results.

\textbf{Heavy Occlusion: }Fig.~\ref{fig:trackingResult} (a) and (b) show four challenging sequences with heavy occlusion.
In \emph{Faceocc1} and \emph{Faceocc2}, the targets undergo
with heavy occlusion and in-plane rotation, it can be seen that our method outperforms the other tracking algorithms.
\emph{Freeman4} and \emph{David3} demonstrate that the proposed
method can capture the accurate location of objects in terms of position, and scale when the target undergoes severe occlusion (\eg, \emph{Freeman4} \#0144 and \emph{David3} \#0085). However, IVT, $L_1$APG, MIL, SP, SCM, ASLA,
TLD, SPOT, FOT, SST, MTT, and Struck methods drift away from the target object when occlusion occurs. For these four sequences, the IVT method performs poorly since conventional PCA is not robust to occlusions. Although $L_1$APG and SP utilize sparsity to model outliers, it is observed that their occlusion detection are not stable when drastic change of appearance happens. In contrast, our method is robust to heavy occlusion. This is because our combination of $L_0$ and $L_1$ regularized appearance model can exactly reconstruct the object.
%and the dictionary reinitialization can capture the intrinsic low-rank features from corrupted observations.
\textbf{Fast Motion: }Fig.~\ref{fig:trackingResult} (c) show
the sequences \emph{Boy} and \emph{Jumping} with fast motion. It is difficult to
predict the locations of the tracked objects when they undergo
abrupt motion.
In \emph{Boy}, the captured images are blurred seriously, but Struck and our method track the target faithfully throughout the images. IVT, MTT, ALSA, SCM and SST methods drift away seriously.
We note that most of the other trackers have drift problem due to the abrupt motion in sequence~\emph{Jumping}. In contrast, the SST and our method successfully track the target for whole video.

%We note that SP performs better in sequence~\emph{Jumping}. However,
%the linear coefficients of this method is obtained by the least square method, which is not able to capture the accurate candidate.

\textbf{Drastic Pose, Scale and Illumination Changes: }In Fig.~\ref{fig:trackingResult} (d) and (e), we test five challenging
sequences with drastic pose, scale and illumination change.
\emph{Fish} and \emph{Tiger1} chips contain significant illumination variation. We can see that the $L_1$APG, MTT, and MIL methods are less effective in these cases (\eg, \emph{Fish} \#0305 and \emph{Tiger1} \#0240).
In \emph{Singer2} and \emph{Jogging-2}, other trackers drift away when objects under variable illumination, and pose variation (\eg, \emph{Singer2} \#0110 and \emph{Jogging-2} \#0100 ), however, our method still performs well. Our method also achieves good performance in \emph{CarScale} with scale variation (\eg, \emph{CarScale} \#0204).
%For these five sequences, the performance of IVT is bad because it is sensitive to drastic change of appearance.
For subspace-based approaches, they may fail to update the appearance model as the calculation of coefficients in their models may have redundant background features. Our method can successfully adapt to variable drastic changes since the combination of sparse coding and sparse counting is not merely stable but also applicable to obtain the intrinsic features of the subspace.

\textbf{Background Clutters: }{Fig.~\ref{fig:trackingResult} (f)} demonstrates the tracking
results in \emph{Deer}, \emph{Baskerball}, and \emph{Football} with background clutter.
\emph{Baskerball} is a difficult sequence because it contains cluttered background, illumination change, heavy occlusion and non-rigid pose variation. Unless our tracker, none of the compared algorithms can work well on it(\eg, \emph{Baskerball} \#0486 and \#0614).
%However, the performance of our method is still relatively well on such kind of videos.
As shown in \emph{Deer} and \emph{Football}, our tracker performs relatively well (\eg, \emph{Deer} \#0031 and \emph{Football} \#304) as it has excluded background clutters in the sparse errors, but TLD, FOT, and MIL fail.
%For these sequences, IVT, SP fails because subspace they learn from the original observations contains cluttered background. In contrast,
%Our method is robust to background clutter because our appearance model is obtained from usefull features which has excluded background clutters in the sparse errors.
\begin{figure}[h]
\begin{center}
\begin{tabular}{c}
\includegraphics[width=0.49\textwidth,keepaspectratio]{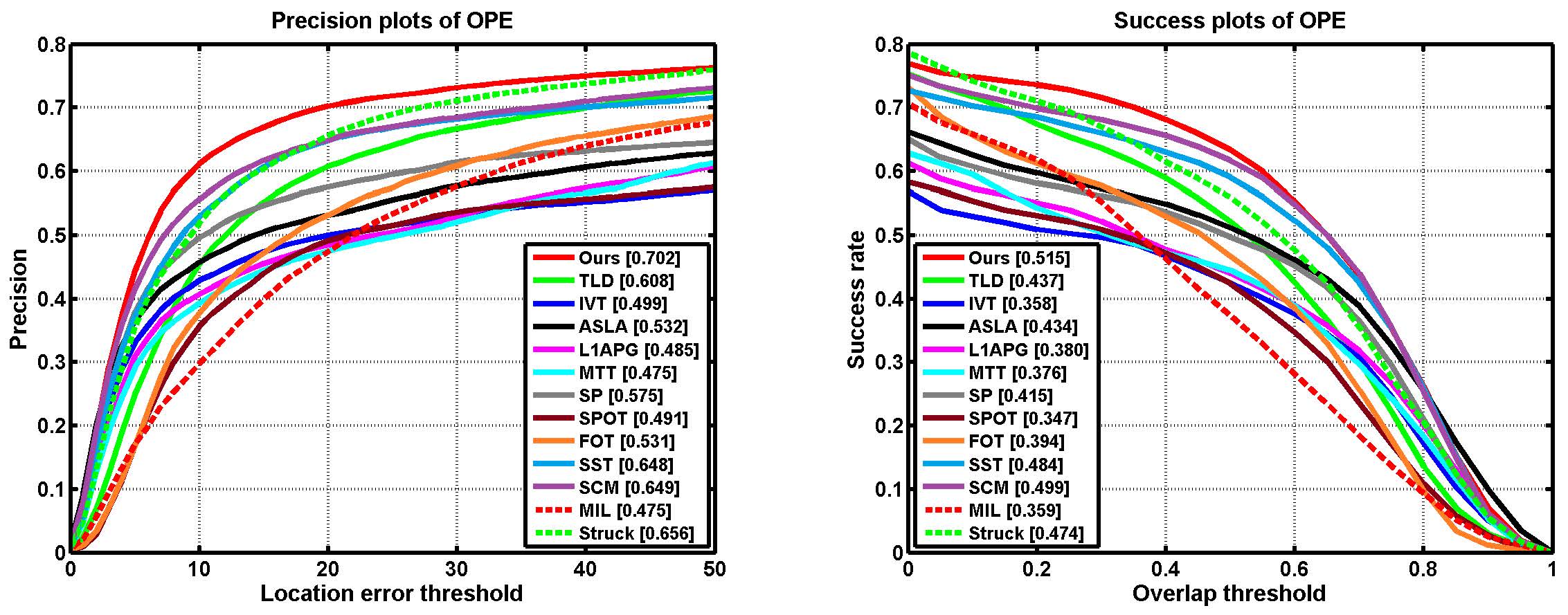}
\end{tabular}
\end{center}
\caption
{ \label{fig: precision}
{\bf Precision and success plots over all the 50 sequences.} The mean precision scores are reported in the legends. }
\end{figure}
\subsection*{Quantitative Evaluation}

\begin{table*}[htbp]
\centering
\footnotesize
\caption{Average Overlap Rate (in pixels) and average frame per second (FPS). The best and the second results are shown in \textcolor[rgb]{1.00,0.00,0.00}{\textbf{BOLD}} fonts and \textcolor[rgb]{0.00,0.00,1.00}{\textbf{BOLD}} fonts, respectively.} \label{tab:ORE} %\vskip
%\begin{tabular}{|@{}r@{}|@{}r@{}|@{}r@{}|@{}r@{}|@{}r@{}|@{}r@{}|@{}r@{}|@{}r@{}|@{}r@{}|@{}r@{}|@{}r@{}|@{}r@{}|@{}r@{}|@{}r@{}|}
\begin{tabular}{|c|c|c|c|c|c|c|c|c|c|c|c|c|c|}
\hline
&{TLD}  &{IVT}  &{ASLA} &{$L_1$APG} &{MTT} &{SP} &{SPOT}&{FOT}&{SST}&{SCM} &{MIL} &{Struck} &{Ours} \\ \hline
Faceocc1  &0.58&0.73&0.32&0.76&0.70&0.79&0.74&0.60&0.79&\textcolor[rgb]{0.00,0.00,1.00}{\textbf{0.79}}&0.60&0.73&\textcolor[rgb]{1.00,0.00,0.00}{\textbf{0.80}}\\ \hline
Faceocc2  &0.62&0.73&0.65&0.69&\textcolor[rgb]{0.00,0.00,1.00}{\textbf{0.75}} &0.59&0.69&0.64&0.63&0.73&0.67&\textcolor[rgb]{1.00,0.00,0.00}{\textbf{0.79}}&0.69\\ \hline
Freeman4  &0.22&0.15&0.13&0.34&0.22&0.17&0.01&0.11&0.18&\textcolor[rgb]{0.00,0.00,1.00}{\textbf{0.26}}&0.05&0.17&\textcolor[rgb]{1.00,0.00,0.00}{\textbf{0.41}}\\ \hline
David3    &0.10&0.48&0.43&0.38&0.10&0.46&\textcolor[rgb]{1.00,0.00,0.00}{\textbf{0.77}}&0.41&0.30&0.41&0.54&0.29&\textcolor[rgb]{0.00,0.00,1.00}{\textbf{0.73}}\\ \hline
Boy       &0.66&0.26&0.37&\textcolor[rgb]{0.00,0.00,1.00}{\textbf{0.73}}&0.50&0.36&0.57&0.64&0.36&0.38&0.49&0.76&\textcolor[rgb]{1.00,0.00,0.00}{\textbf{0.81}}\\ \hline
Jumping   &0.66&0.12&0.23&0.15&0.10&\textcolor[rgb]{0.00,0.00,1.00}{\textbf{0.70}}&0.01&0.20&0.16&0.62&0.12&0.52&\textcolor[rgb]{1.00,0.00,0.00}{\textbf{0.71}}\\ \hline
Fish      &0.81&0.77&0.85&0.34&0.16&0.83&0.83&0.78&\textcolor[rgb]{0.00,0.00,1.00}{\textbf{0.86}}&0.75&0.45&0.85&\textcolor[rgb]{1.00,0.00,0.00}{\textbf{0.87}}\\ \hline
Tiger1    &0.38&0.10&0.29&0.31&0.26&0.10&\textcolor[rgb]{1.00,0.00,0.00}{\textbf{0.70}}&0.19&0.16&0.16&0.12&0.15&\textcolor[rgb]{0.00,0.00,1.00}{\textbf{0.61}}\\ \hline
Singer2   &0.22&0.04&0.04&0.04&0.04&0.04&\textcolor[rgb]{1.00,0.00,0.00}{\textbf{0.75}}&0.21&0.04&0.17&0.51&0.04&\textcolor[rgb]{0.00,0.00,1.00}{\textbf{0.62}}\\ \hline
Jogging-2 &0.66&0.14&0.14&0.15&0.13&\textcolor[rgb]{0.00,0.00,1.00}{\textbf{0.73}}&0.20&0.12&0.12&0.73&0.14&0.20&\textcolor[rgb]{1.00,0.00,0.00}{\textbf{0.74}}\\ \hline
CarScale  &0.45&\textcolor[rgb]{0.00,0.00,1.00}{\textbf{0.63}}&0.61&0.50&0.49&0.60&0.01&0.35&0.55&0.59&0.41&0.41&\textcolor[rgb]{1.00,0.00,0.00}{\textbf{0.81}}\\ \hline
Deer      &0.60&0.03&0.03&0.60&0.61&0.72&0.72&016&0.62&0.07&0.12&\textcolor[rgb]{0.00,0.00,1.00}{\textbf{0.74}}&\textcolor[rgb]{1.00,0.00,0.00}{\textbf{0.82}}\\ \hline
Basketball&0.02&0.11&0.39&0.23&0.19&0.23&0.01&0.17&0.20&\textcolor[rgb]{0.00,0.00,1.00}{\textbf{0.46}}&0.22&0.20&\textcolor[rgb]{1.00,0.00,0.00}{\textbf{0.63}}\\ \hline
Football  &0.49&0.56&0.53&0.55&0.58&\textcolor[rgb]{1.00,0.00,0.00}{\textbf{0.69}}&0.01&0.55&0.40&0.49&0.59&0.53&\textcolor[rgb]{0.00,0.00,1.00}{\textbf{0.59}}\\ \hline
Average   &0.46&0.34&0.36&0.41&0.34&\textcolor[rgb]{0.00,0.00,1.00}{\textbf{0.50}}&0.43&0.37&0.39& 0.44&0.39&0.41&\textcolor[rgb]{1.00,0.00,0.00}{\textbf{0.70}}\\ \hline \hline
FPS    &21.74&\textcolor[rgb]{0.00,0.00,1.00}{\textbf{27.83}}&7.48&2.47&0.99&2.35&--&\textcolor[rgb]{1.00,0.00,0.00}{\textbf{376.48}}&2.12&
0.37&28.06&10.01&4.27\\ \hline
\end{tabular}
\end{table*}

We use two metrics to evaluate the proposed algorithm with other state-of-the-art methods.
The first metric is the center location error measured with manually labeled ground truth data.
The second one is the overlap rate, i.e., $score = \frac{area(R_T\bigcap R_G)}{area(R_T\bigcup R_G)}$, where $R_T$ is the tracking bounding box and $R_G$ is the ground truth bounding box. The larger average scores mean more accurate results.

\begin{figure*}[htb!]
\begin{center}
\begin{tabular}{c}
\includegraphics[width=0.98\textwidth,keepaspectratio]{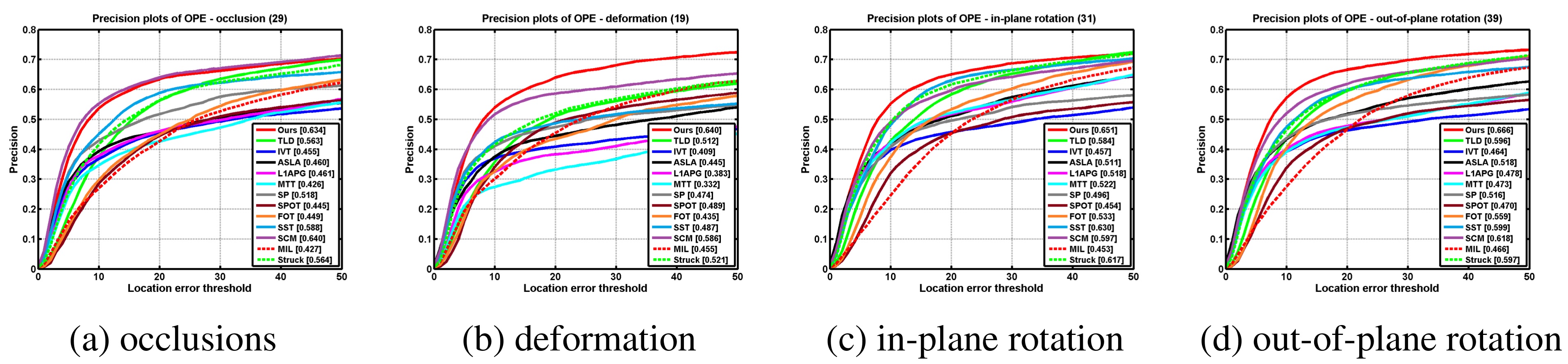}
\end{tabular}
\end{center}
\caption
{ \label{fig: center error}
{\bf The plots of OPE with attributes based on the precision metric. }}
\end{figure*}

\begin{figure*}[htb!]
\begin{center}
\begin{tabular}{c}
\includegraphics[width=0.98\textwidth,keepaspectratio]{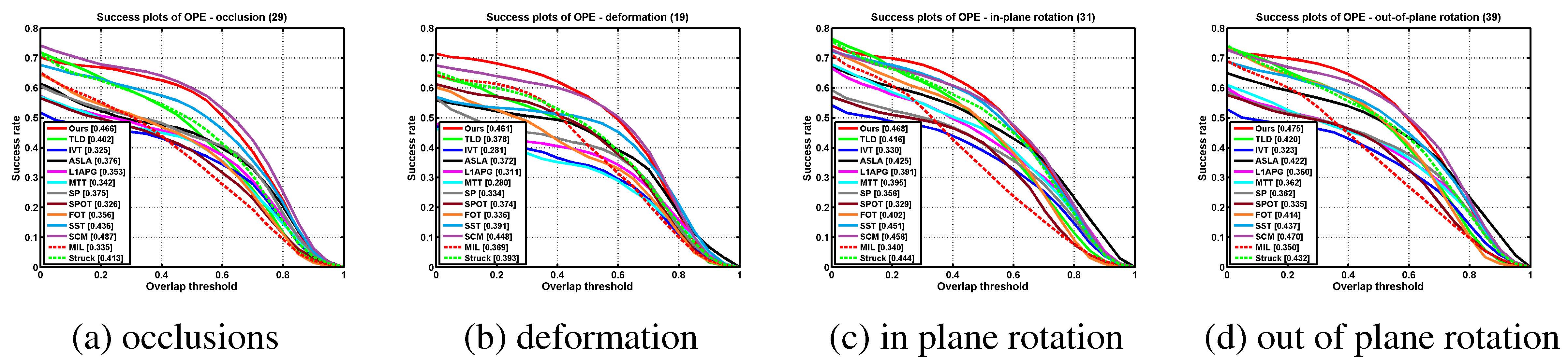}
\end{tabular}
\end{center}
\caption
{ \label{fig: overlap rate}
{\bf The plots of OPE with attributes using the success rate metric. }}
\end{figure*}

Table \ref{tab:ORE} shows the average overlap rates. Table \ref{tab:CEE} reports the average center location errors (in pixels) where a smaller average error means a more accurate result. As can be seen from the table, the most sequences generated by our method have lower average error and higher overlap rate values. We provide the precision and success plots in Fig.~\ref{fig: precision} to evaluate our performance over all the 50 sequences. The evaluation parameters are set as default in~\cite{survey2/cvpr/Yang}.
We note that the our algorithm performs well for the videos with occlusion, deformation,
in plane rotation, and out of plane rotation based on the precision
metric and the success rate metric as shown in Fig.~\ref{fig: center error}
and Fig.~\ref{fig: overlap rate} respectively. Both table and figures show that our method achieves favorable performance against other state-of-the-art methods.

To further compare the running time of four subspace-based tracking algorithms (i.e. IVT, $L_1$APG, SP and our method), we calculated the average Frames Per Second (FPS) for $32\times32$ image patch (see the last row of Table~\ref{tab:ORE}). For $L_1$APG, we reported FPS for its APG acceleration.
It can be seen that IVT is quite faster than other trackers as its computation only involves matrix-vector multiplication. Both SP and our method are faster than $L_1$APG. It is also observed that our method is much faster than SP. This is due to the different choices of the optimization scheme. SP adopts a naive altering minimization strategy, in contrast, our method is efficiently solved by APG.
\begin{table*}[htbp]
\centering
\footnotesize
\caption{ Average Center Location Error(in pixels) and average frame per second (FPS). The best and the second results are shown in \textcolor[rgb]{1.00,0.00,0.00}{\textbf{BOLD}} fonts and \textcolor[rgb]{0.00,0.00,1.00}{\textbf{BOLD}} fonts, respectively.} \label{tab:CEE} %\vskip
%\begin{tabular}{|@{}r@{}|@{}r@{}|@{}r@{}|@{}r@{}|@{}r@{}|@{}r@{}|@{}r@{}|@{}r@{}|@{}r@{}|@{}r@{}|@{}r@{}|@{}r@{}|@{}r@{}|@{}r@{}|}
\begin{tabular}{|c|c|c|c|c|c|c|c|c|c|c|c|c|c|}
\hline
&{TLD}  &{IVT}  &{ASLA} &{$L_1$APG} &{MTT} &{SP} &{SPOT}&{FOT}&{SST}&{SCM} &{MIL} &{Struck} &{Ours} \\ \hline
%&{~~TLD}  &{~~IVT}  &{~ASLA} &{$L_1$APG} &{~~MTT} &{~~~~SP} &{~SPOT}&{~~FOT}&{~~SST}&{~~SCM} &{~~MIL} &{Struck} &{~~Ours} \\ \hline
Faceocc1  &27.37&18.42&78.06&17.33&21.00&14.14&17.17&29.00&13.00&\textcolor[rgb]{0.00,0.00,1.00}{\textbf{13.04}}&29.86&18.78&\textcolor[rgb]{1.00,0.00,0.00}{\textbf{12.88}}\\ \hline
Faceocc2  &12.28 &7.42 &19.35 &12.76 &9.836&10.43&11.78&11.94&12.82&\textcolor[rgb]{0.00,0.00,1.00}{\textbf{5.96}} &9.02 &13.60&\textcolor[rgb]{1.00,0.00,0.00}{\textbf{5.50}}\\ \hline
Freeman4  &39.18&43.04&70.24&\textcolor[rgb]{0.00,0.00,1.00}{\textbf{22.12}}&23.55&79.66&108.70&54.66&56.20&56.20&62.07&48.70&\textcolor[rgb]{1.00,0.00,0.00}{\textbf{10.39}}\\ \hline
David3    &208.00&51.95&87.76&90.00&341.33&\textcolor[rgb]{0.00,0.00,1.00}{\textbf{8.74}}&6.27&33.40&104.50&73.09&29.68&106.50&\textcolor[rgb]{1.00,0.00,0.00}{\textbf{5.79}}\\ \hline
Boy       &4.49&91.25&106.07&7.03&12.77&58.09&8.93&5.79&66.97&51.02&12.83&\textcolor[rgb]{0.00,0.00,1.00}{\textbf{3.84}}&\textcolor[rgb]{1.00,0.00,0.00}{\textbf{2.57}}\\ \hline
Jumping  &5.94&61.56&46.08&83.75&84.57&\textcolor[rgb]{1.00,0.00,0.00}{\textbf{4.72}}&120.37&19.83&45.70&6.54 &65.89&9.99&\textcolor[rgb]{0.00,0.00,1.00}{\textbf{4.99}}\\ \hline
Fish      &6.54&5.67&3.85&29.43&45.50&3.99&4.52&6.50&\textcolor[rgb]{0.00,0.00,1.00}{\textbf{3.14}}&8.54&24.14&3.40&\textcolor[rgb]{1.00,0.00,0.00}{\textbf{3.08}}\\ \hline
Tiger1    &49.45&106.61&55.87&58.45&64.39&124.36&\textcolor[rgb]{1.00,0.00,0.00}{\textbf{15.93}}&73.49&93.49&93.49&108.93&128.70&\textcolor[rgb]{0.00,0.00,1.00}{\textbf{18.64}}\\ \hline
Singer2   &58.32&175.46&175.28&180.87&209.69&178.39&\textcolor[rgb]{1.00,0.00,0.00}{\textbf{13.73}}&57.62&175.28&113.63&22.53&174.32&\textcolor[rgb]{0.00,0.00,1.00}{\textbf{14.45}}\\ \hline
Jogging-2 &13.56&138.22&169.87&145.85&157.12&\textcolor[rgb]{1.00,0.00,0.00}{\textbf{3.61}}&72.23&169.16&442.77&\textcolor[rgb]{0.00,0.00,1.00}{\textbf{4.15}}&132.99&107.687&5.88\\ \hline
CarScale  &22.60&11.90&24.64&79.78&87.61&\textcolor[rgb]{0.00,0.00,1.00}{\textbf{13.36}}&207.01&106.20&87.05&33.38&33.47&36.43&\textcolor[rgb]{1.00,0.00,0.00}{\textbf{7.66}}\\ \hline
Deer      &30.93&182.69 &160.06&24.19&18.91&6.84&13.95&80.30&13.81&103.54&100.73&\textcolor[rgb]{0.00,0.00,1.00}{\textbf{5.27}}&\textcolor[rgb]{1.00,0.00,0.00}{\textbf{4.59}}\\ \hline
Basketball 	&213.86&107.11&82.64&137.53&106.80&\textcolor[rgb]{0.00,0.00,1.00}{\textbf{39.79}}&169.86&118.02&105.93&52.90&91.92&118.6&\textcolor[rgb]{1.00,0.00,0.00}{\textbf{7.92}}\\ \hline
Football  &14.26&14.34&15.00&15.11&13.67&\textcolor[rgb]{1.00,0.00,0.00}{\textbf{5.22}}&202.03&13.36&17.21&16.30&12.09&17.31&\textcolor[rgb]{0.00,0.00,1.00}{\textbf{7.28}}\\ \hline
Average   &50.48&72.54&78.20&64.58&85.48&\textcolor[rgb]{0.00,0.00,1.00}{\textbf{39.38}}&69.46&55.66&88.42
&48.26&48.92&49.17&\textcolor[rgb]{1.00,0.00,0.00}{\textbf{7.97}}\\ \hline \hline
FPS    &21.74&\textcolor[rgb]{0.00,0.00,1.00}{\textbf{27.83}}&7.48&2.47&0.99&2.35&--&\textcolor[rgb]{1.00,0.00,0.00}{\textbf{376.48}}&2.12&
0.37&28.06&10.01&4.27\\ \hline
\end{tabular}
\end{table*}


\section*{Conclusion}
In this paper, we propose sparse coding and counting method under Bayesian framwork for robust visual tracking. The proposed method combines $L_0$ regularization and $L_1$ regularized sparse representation in a unique one, therefore, it has better ability to sparsely represent an object and the reconstruction result are also better.
Besides, to solve the proposed model, we develop a fast and efficient APG algorithm.
Moreover, the closed solution of the combination of $L_0$ norm and $L_1$ norm regularization is provided.
Extensive experiments testify to the superiority of our method over state-of-the-art methods, both qualitatively and quantitatively.

\section*{Acknowledgment}
This work is partially supported by the National Natural Science Foundation of China (Nos. 61300086, 61432003, 61301270, 61173103, 91230103), the Fundamental Research Funds for the Central Universities (DUT15QY15), the Open Project Program of the State Key Laboratory of CAD\&CG, Zhejiang University, Zhejiang, China (No. A1404), and National Science and Technology Major Project~(Nos. 2013ZX04005-021, 2014ZX001011).
\section*{Appendix: Proof of Lemma~\ref{lemma:1}}
\label{appendix-1}
\begin{proof}
First, we denote $E(x)=\frac{1}{2}(x-y)^2+\delta|x|+\eta|x|_0$. It is obvious that if $x=0$, then $E(0)=\frac{1}{2}y^2$. Then we need to discuss the case that $x\neq0$:
\begin{enumerate}
\item if $x>0$, then $E(x)=\frac{1}{2}(x-y)^2+\delta x+\eta$. Writing its K.K.T condition, we get $x=y-\delta$, and the objective value is $E(y-\delta)=-\frac{1}{2}\delta^2+\delta y+\eta$.
\item if $x<0$, then $E(x)=\frac{1}{2}(x-y)^2-\delta x+\eta$. It is easy to get $x=y+\delta$, and the objective value is $E(y+\delta)=-\frac{1}{2}\delta^2-\delta y+\eta$.
\end{enumerate}
Then, we need to compare these three cases, if $E(0)>E(x-\delta)$, we have $(\delta-y)^2>2\eta$. Combining with $x=y-\delta>0$, we have $y > \delta+\sqrt{2\eta}$. Similarly, if $E(0)>E(x+\delta)$, then we have $y < -\delta-\sqrt{2\eta}$. And $x =0$, otherwise.
\end{proof}

\bibliographystyle{model1b-num-names}
\bibliography{refs}

%% not want to use model1b-num-names.bst.





% \end{thebibliography}

\end{document}